\documentclass[sigconf]{acmart}

\AtBeginDocument{%
  \providecommand\BibTeX{{%
    \normalfont B\kern-0.5em{\scshape i\kern-0.25em b}\kern-0.8em\TeX}}}

\copyrightyear{2022}
\acmYear{2022}

\setcopyright{acmlicensed}
\acmConference[WWW '22]{Proceedings of the ACM Web Conference 2022}{April 25--29, 2022}{Virtual Event, Lyon, France}
\acmBooktitle{Proceedings of the ACM Web Conference 2022 (WWW '22), April 25--29, 2022, Virtual Event, Lyon, France}
\acmPrice{15.00}
\acmDOI{10.1145/3485447.3512260}
\acmISBN{978-1-4503-9096-5/22/04}

\usepackage{xcolor}

\usepackage[belowskip=0pt,aboveskip=0pt]{caption}

\usepackage{multirow}

\begin{document}

\title{On Explaining Multimodal Hateful Meme Detection Models}


\author{Ming Shan Hee}
\affiliation{%
  \institution{Singapore University of \\Technology and Design}
  \city{Singapore}
  \country{Singapore}
}
\email{mingshan_hee@mymail.sutd.edu.sg}

\author{Roy Ka-Wei Lee}
\affiliation{%
  \institution{Singapore University of \\Technology and Design}
  \city{Singapore}
  \country{Singapore}
}
\email{roy_lee@sutd.edu.sg}

\author{Wen-Haw Chong}
\affiliation{%
  \institution{Singapore Management University}
  \city{Singapore}
  \country{Singapore}
}
\email{whchong.2013@phdis.smu.edu.sg}

\renewcommand{\shortauthors}{Ming Shan Hee, Roy Ka-Wei Lee, \& Wen-Haw Chong}

\begin{abstract}
Hateful meme detection is a new multimodal task that has gained significant traction in academic and industry research communities. Recently, researchers have applied pre-trained visual-linguistic models to perform the multimodal classification task, and some of these solutions have yielded promising results. However, what these visual-linguistic models learn for the hateful meme classification task remains unclear. For instance, it is unclear if these models are able to capture the derogatory or slurs references in multimodality (i.e., image and text) of the hateful memes. To fill this research gap, this paper propose three research questions to improve our understanding of these visual-linguistic models performing the hateful meme classification task. We found that the image modality contributes more to the hateful meme classification task, and the visual-linguistic models are able to perform visual-text slurs grounding to a certain extent. Our error analysis also shows that the visual-linguistic models have acquired biases, which resulted in false-positive predictions.

\end{abstract}




\maketitle
{\color{red} \textbf{Disclaimer}: \textit{This paper contains violent and discriminatory content that may be disturbing to some readers.}} 
\section{Introduction}







\textbf{Motivation.} Internet memes, which are often presented as images with accompanying text, are increasingly abused to spread hatred under the guise of humor~\cite{kiela2020hateful,gomez2020exploring,pramanick2021detecting}. To fight against the proliferation of hateful memes, Facebook has recently released a large hateful meme dataset and crowdsourced hateful meme classification solutions~\cite{kiela2020hateful}. The research community has responded enthusiastically as many promising hateful meme classification methods have been proposed~\cite{lippe2020multimodal,zhu2020enhance,velioglu2020detecting,zhou2021multimodal,lee2021disentangling}. Among the proposed solutions, a popular line of approaches is to apply pre-trained visual-linguistic models such as VisualBERT~\cite{li2019visualbert} and VilBERT~\cite{Lu2019ViLBERTPT} to perform the hateful meme classification tasks. These methods have yielded promising results. However, what these visual-linguistic models learn for the hateful meme classification task remains unclear. 

\textbf{Research Objectives.} Understanding visual-linguistic models is an emerging research area that has garnered much attention from the multimodal research community~\cite{li2020does,cao2020behind,frank2021vision,parcalabescu2021seeing}. Inspired by works that explored the internal behaviors of pre-trained language models~\cite{clark2019does}, Li et al.~\cite{li2020does} conducted a quantitative study on whether visual-linguistic models acquire semantic grounding ability during pre-training without explicit supervision. Frank et al.~\cite{frank2021vision} proposed a diagnostic framework to assess the extent to which the visual-linguistic models integrate cross-model information. This paper aims to contribute to the existing literature on visual-linguistic model understanding by applying some of these techniques to investigate how visual-linguistic models understand hateful memes. To the best of our knowledge, this is the first paper that attempts to understand what visual-linguistic models actually learn when training for the hateful meme classification task.

\textbf{Contributions.} Our paper proposed three research questions, which improve our understanding of the internal behaviors of visual-linguistic models trained to perform the hateful meme classification task. Through extensive quantitative and qualitative analyses, we show that (i) the visual-linguistic models have accorded higher importance to the visual modality when performing hateful meme classification; (ii) the visual-linguistic models are able to learn the visual-text slurs grounding; (iii) and the models have acquired biases that adversely affected their hateful meme classification performance.


\section{Research Questions}
This paper aims to improve our understanding of visual-linguistic models applied to perform the hateful meme classification task. Working towards this goal, we formulate three research questions to guide our exploration\footnote{Code implementation: https://gitlab.com/bottle\_shop/safe/ExplainHatefulMeme.}.

\textbf{RQ1: Modality Attribution.} A key characteristic of memes is their multimodality nature, where their underlying message is often communicated via a combination of text and visual information. The multimodality characteristic also motivated the application of visual-linguistic models to perform hateful meme classification. However, it is unclear if the text and visual information contributed equally towards the multimodal classification task. Existing studies have attempted to improve the explainability of deep learning models by attributing the prediction of a deep network to its input features~\cite{kokhlikyan2020captum,guidotti2018survey,ancona2018towards,sundararajan2017axiomatic}. For instance, Sundararajan et al.~\cite{sundararajan2017axiomatic} proposed an attribution method called \textit{Integrated Gradients} to score the contribution of input features on deep models' prediction. The researchers applied their model on several images and text deep learning models to demonstrate its ability to explain the prediction results of these models.  Similar studies were also conducted by investigating the attribution of text and visual features in multimodal tasks such as Visual-question-answering (VQA)~\cite{goyal2016towards,goyal2017making,patro2020robust}. We aim to apply the attribution methods to understand how the different modalities input features contribute to the hateful meme classification task.  

\textbf{RQ2: Visual-Text Slurs Grounding.} Hate speech detection tasks have been notoriously challenging due to the ambiguity and variability in natural languages. Adding to the complexity, hate speech also includes unique derogatory terms that are commonly used as insinuations or allegations about members of a specific group. For example, the word \textit{"Dishwasher"} has been associated with females due to the traditional gender ideology where married women would become housewives and do house chores. In doing so, it undermines women's gender equality rights and objectifies them as mere tools. The use of derogatory terms amplifies further in hateful memes. With the additional visual information, the subtle allegations in the derogatory terms are communicated through contextual cues in both modalities. Existing works have investigated the semantic grounding capabilities of visual-linguistic models in VQA and image captioning tasks~\cite{li2020does,frank2021vision}. We aim to extend these studies to understand the visual-linguistic models' ability to perform visual-text grounding for derogatory terms and slurs used in the hateful memes.    

\textbf{RQ3: Bias and Error Analysis.} Data and model biases in text-based hate speech detection tasks have been are widely researched~\cite{badjatiya2019stereotypical,davidson2019racial,kennedy2020contextualizing,xia2020demoting,zhou2021challenges}. For instance, Kennedy et al.~\cite{kennedy2020contextualizing} conducted a study to analyze group identifier biases in hate speech detection models. The researchers found that existing text-based hate speech classifiers are over-sensitivity to group identifiers like ``\textit{Muslim}'', ``\textit{gay}'', and ``\textit{black}''. We aim to extend these studies to conduct a preliminary analysis on the biases in hateful meme classification models. Specifically, we will examine the group identifier biases in both text and visual modalities of the wrongly classified memes.

\section{Experiments}
\subsection{Experiment Settings}

\subsubsection{Dataset}

\begin{table}[t]
  \caption{Distribution of Facebook hateful meme dataset}
  \label{tab:dataset}
  \begin{tabular}{c|c|c|c}
    \hline
    \multicolumn{2}{c|}{Train} &\multicolumn{2}{c}{Test} \\
    \hline\hline
    Hate & Non-hate &Hate &Non-hate \\ 
    \hline
     5,493&3,007 &246&254  \\
    \hline
\end{tabular}
\end{table}

The Facebook hateful meme dataset~\cite{kiela2020hateful}, which was constructed and released by Facebook as part of a challenge to crowd-source multimodal hateful meme classification solutions, is a popular dataset used in many research studies. Therefore, we utilize this dataset in our experiments. The dataset contains $10K$ memes with binary labels (i.e., hateful or non-hateful). As we do not have labels of the memes in the test split, we utilize the \textit{dev-seen} split as the \textit{test} set. Table~\ref{tab:dataset} outlines the distributions of the dataset.


\subsubsection{Models}

VilBERT \cite{Lu2019ViLBERTPT} and VisualBERT \cite{li2019visualbert} are amongst the state-of-the-art visual-linguistic models often used for various multimodal tasks. The two models were also applied as baselines to evaluate the released Facebook hateful meme dataset~\cite{kiela2020hateful}. Both models use pre-trained text from BERT \cite{devlin2018bert} and image features from $fc6$ layer of Faster-RCNN \cite{ren2015faster} with ResNeXt-152 as its backbone \cite{xie2017aggregated}. These models can also be trained on multimodal objectives as an intermediate step before fine-tuning them for the multimodal hateful meme classification task. For our experiments, we train the \textbf{VilBERT} and \textbf{VisualBERT} on the hateful meme classification task. We also included two multimodally pre-trained versions of these models and fine-tuned them for the hateful meme classification task. Specifically, we include VilBERT on Conceptual Captions \cite{sharma2018conceptual} (\textbf{VilBERT CC}), and VisualBERT on Microsoft's Common Objects in Context \cite{lin2014microsoft} (\textbf{VisualBERT COCO}).  We trained the models using the Facebook MMF framework~\cite{singh2020mmf} and adopt the hyperparameters specified in \cite{kiela2020hateful}, as the researchers have already performed grid search on numerous hyperparameters.






\begin{table}[t]
\caption{Average gradients from text and visual inputs in various models}
\label{table:gradientcontribution}
\begin{tabular}{l|c|c|c|c}
    \hline
    \multirow{2}{*}{Model}& \multicolumn{2}{|c|}{Text Input} & \multicolumn{2}{|c}{Visual Input} \\ \cline{2-5}
     & Avg & Std. Dev & Avg & Std. Dev \\ \hline\hline
    VilBERT         & 3.183 & 0.900 & 4.096 & 0.774  \\\hline
    VisualBERT      & 3.006 & 0.804 & 7.705 & 2.180 \\\hline 
    VilBERT CC      & 3.130 & 0.882 & 4.145 & 0.730 \\\hline
    VisualBERT COCO & 3.112 & 0.843 & 6.444 & 0.976 \\\hline
\end{tabular}
\end{table}

\begin{figure*}[t]
    \centering
    \includegraphics[width=1\textwidth]{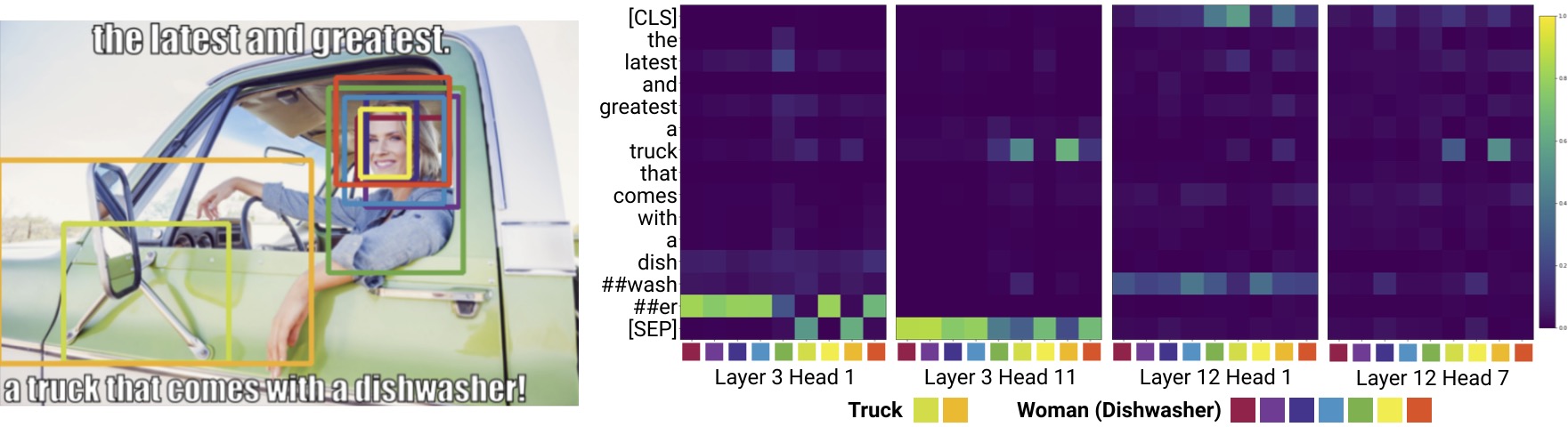}
    \caption{Attention heads in VisualBERT where the text-visual alignments attend to the word "dishwasher" and "truck".}
    \Description[an image depicting a female driver driving a truck and the attention weights obtained from Transformer architecture]{the attention weight depicts strong visual-text alignments between visual objects and certain words such as "truck" and "dishwasher"}
    \label{fig:slurs-grounding-dishwasher}
\end{figure*}

\begin{figure*}[t]
    \centering
    \includegraphics[width=1\textwidth]{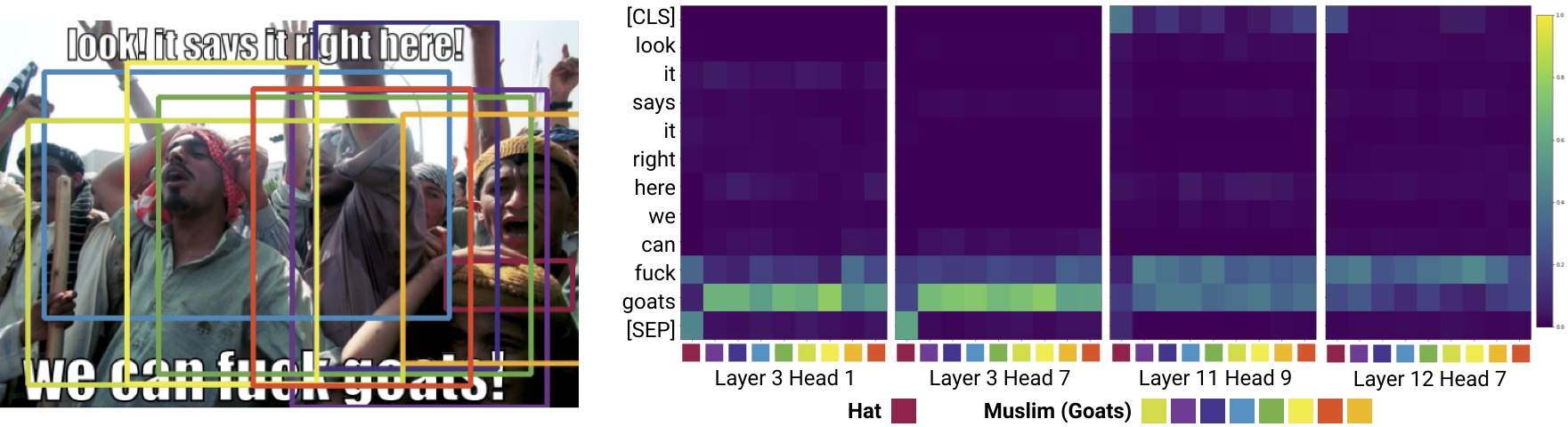}
    \caption{Attention heads in VisualBERT where the text-visual alignments attend to the word "goat" and "f*ck"}
    \Description[an image depicting a group of Muslim gathering and the attention weights obtained from Transformer architecture]{the attention weight depicts strong visual-text alignments between visual objects and certain words such as "goat" and "fu*k"}
    \label{fig:slurs-grounding-goat}
\end{figure*}

\subsection{Modality Attribution}
Inspired by gradient-based researches that use gradients as feature importance \cite{selvaraju2017grad, simonyan2013deep}, we use the gradients to represent the contribution of each modality towards making the model's decision. We reasoned that gradients signify the weightage each input feature has towards making the model's prediction. Hence, we equate the summation of gradients for each input type as the contribution of each modality. Specifically, we first obtain the gradients for the inputs in each modality through backpropagation. Subsequently, we normalize the gradients by their magnitude. Through normalization, we place the gradients for the text and visual modality onto the same unit space and enable us to compare their relative contribution. We attribute the summation of these normalized gradients as the contribution of each modality. Finally, we compute the average and standard deviation across the 500 samples in the test data. We noted empty tokens are present in the text inputs but not in the visual inputs, as the text inputs have various lengths. As these empty tokens do not contain any meaning, we argue that their gradients should not attribute to the contribution of each modality and do not consider these gradients in our analysis.

Table \ref{table:gradientcontribution} shows the average and standard deviation for the text and visual modality gradients for various hateful meme classification models. We observe that the visual modality consistently has a slightly higher average gradient than the text modality, suggesting that the visual modality contributes more towards the model's prediction. We also noted that the standard deviation for both modalities is relatively low across most models, suggesting that there is little variation in the contribution of each modality across all samples.

The observation that the visual modality contributes more in the hateful meme classification results concurs with Frank et al.~\cite{frank2021vision}'s findings where proposed vision-and-language models tend to attend to visual information more than text information. A possible reason for this observation could be the ratio of text features to image features. We note that internet memes generally have relatively short text. For instance, the samples in the validation dataset have an average number of 14 words and a maximum number of 54 words. Conversely, the image features always comprise the top 100 image regions extracted from existing object detection techniques. Hence, the number of text features is always lower than the number of image features across the samples, and the models may have leverage on the modality with more features.

\subsection{Visual-Text Slurs Grounding} \label{subsec:semantic-grounding}


The visual-linguistic models offer a simplistic solution to learn interaction across modalities. It uses a stack of Transformer layers that implicitly aligns the text input and visual input with self-attention. Recent works have also established that input alignments within VisualBERT's attention weights often capture intricate associations in the Transformers' architecture (e.g., visual-text entity grounding, visual-text syntactic grounding, etc.) \cite{li2020does}. Extending from these studies, we should observe visual-text alignments for slurs within the attention weights after fine-tuning the visual-linguistic models on the hateful meme classification task.

As our investigation aims to understand the visual-text grounding in the hateful meme classification tasks, we visualize the visual-text alignments within the models' attention weights for evidence of slurs grounding. Specifically, we display the bounding region on the image for each visual feature. To obtain the bounding region for each visual feature, we used the predicted coordinates from the final layer of the Faster-RCNN as they originate from the model's intermediate layer fc6. However, visualizing 100 bounding regions would clutter the image and would be impractical to make any valuable observation. Therefore, we choose to visualize the top 9 features ranked by their contribution to the models' decisions. 

There are many slurs embedded in the hateful memes. We select and examine two common slurs, ``\textit{dishwasher}'' and ``goat-f*cker'', that target female and Muslim communities, respectively. Specifically, we search the hateful memes that contain the words ``\textit{dishwasher}'', ``\textit{goat}'', and ``goat-f*cker''. From the retrieved memes, we examine the visual bounding regions that have attention weight aligned to the selected keywords. We then show the 4 heads of the attention layers that best demonstrate visual-text alignments for slurs and entities. The visual-text alignment will reveal the models' multimodal understanding of the slurs embedded in the memes. 

The word ``\textit{dishwasher}'' is a sexist slur commonly found in hate speech targeted at the female gender, used to stigmatize women as housewives. In Figure \ref{fig:slurs-grounding-dishwasher}, the meme describes a scenario where one undermines women's gender equality rights and addressed the woman as ``\textit{dishwasher}''. We observed that the model implicitly forms alignments between the subword "\#\#er" for \textit{dishwasher} and the seven image segments containing the woman in lower layers. The model displays residual visual-text alignments for the slur using a different subword "\#\#wash" by the end of the computation (i.e., layer 12), demonstrating the strong presence of visual-text alignment for image of woman and the slur ``\textit{dishwasher}''.

\begin{figure}[t]
    \centering
    \includegraphics[width=0.44\textwidth]{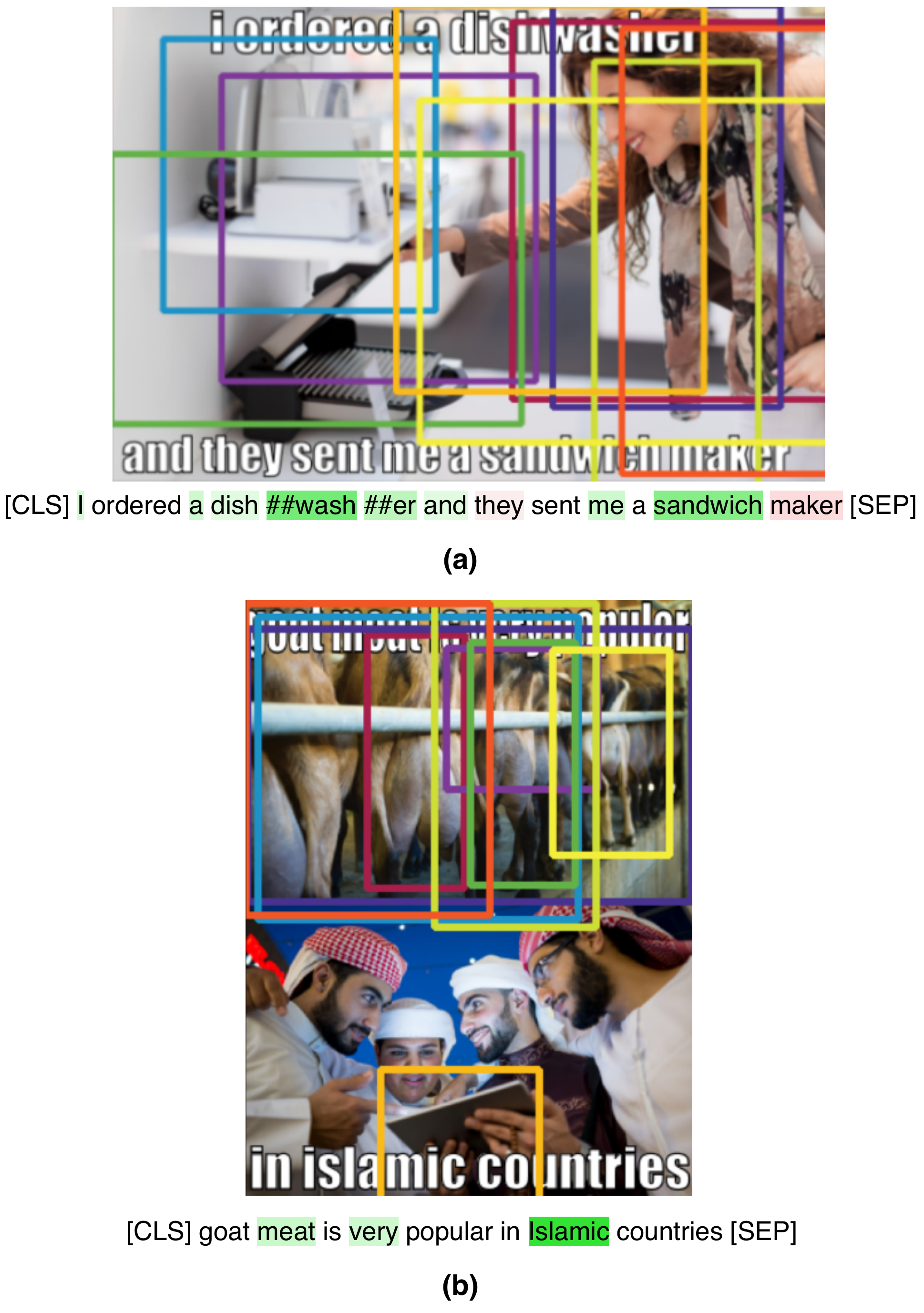}
    \Description[a top image containing female opening a sandwich maker and a bottom image containing muslims with goats]{The primary text contribution of the top image comes from two subwords resulting from "dishwasher" and "sandwich". Whereas, the primary text contribution from bottom image come from the word "meat", "very" and "islamic"}
    \caption{Two examples of non-hateful memes wrongly classified by VisualBERT}
    \label{fig:error-analysis}
\end{figure}

The phrases ``\textit{goat-humper}'' and ``\textit{goat-f*cker}'' is one of the many offensive slurs commonly targeted at the Muslim religion, accusing Muslim men of having sexual relationships with goats (i.e., an act of bestiality). Figure \ref{fig:slurs-grounding-goat} shows a meme that suggests the Muslim men have sexual relationships with goats by associating the text ``\textit{we can f*ck goats}'' with Muslim men in the image. While the phrase ``\textit{goat-humper}'' and ``\textit{goat f*cker}'' does not appear in this meme, we could infer the underlying allegations and identify the relevant keywords. We observed that the word ``\textit{f*ck}'' and the word ``\textit{goat}'' assigns significant attention weights to the bounding regions containing Muslim men in the early layers. The model displays residual visual-text alignment for the word ``\textit{f*ck}'' and the word ``\textit{goat}'' in the later layers (i.e., layer 11 and 12), albeit more attention weights are assigned to the word ``\textit{f*ck}''. Combining the observations, the model demonstrates presence of visual-text alignment for the relevant keywords that suggest the slurs ``\textit{goat-humper}'' and ``\textit{goat-f*cker}''. 


We also observed that the early layers demonstrate a cleaner alignment to the slur terms. For example, in Figure \ref{fig:slurs-grounding-dishwasher}, the unrelated visual segments to the subword "\#\#er" and the word "truck" are aligned to the separator token.  A similar observation can be made for the unrelated visual segments to the word "goat" in Figure \ref{fig:slurs-grounding-goat}. Based on recent researches, these alignments to the separator tokens can be seen as no-operation (no-op) as they do not substantially impact the model's output \cite{clark2019does, kobayashi2020attention}.

\subsection{Bias and Error Analysis}
To analyze the bias of the hateful meme classification models, we conduct an error analysis on the wrongly predicted non-hateful memes by the models. Similar to the existing works that studied biases in text-based hate speech detection models, we are especially interested in the false positives because it may reveal the features that the models are overly sensitive to when performing hateful meme prediction. Specifically, we conduct this analysis by inspecting the bounding regions for the critical visual features and contributions of individual word tokens for text features. The visualization of bounding regions for the visual features follows the same strategy specified in \ref{subsec:semantic-grounding}. Whereas for the text features, we used Integrated Gradients~\cite{sundararajan2017axiomatic} to visualize the contribution of each word towards making the model's prediction. Specifically, words are highlighted in green and red to indicate their attribution to hateful and non-hateful prediction, and the color intensity represents the level of attribution.



Figure~\ref{fig:error-analysis} shows two examples of non-hateful memes wrongly classified by VisualBERT. In Figure~\ref{fig:error-analysis}(a), we observed that the subword tokens for ``\textit{dishwasher}'' and word token ``\textit{sandwich}'' have a high contribution towards making the model predict the meme as hateful. Upon inspecting the text-visual alignment, these text tokens also assign significant attention weights to the bounding region containing the woman. From the observations, we can infer that the model has likely learned a bias where the presence of keywords such as ``\textit{dishwasher}'' and images of women would render the model to predict the meme as hateful. The bias also exposes a deeper issue with the visual-linguistic hateful meme classification model; even though the models are able to learn the visual-text slurs grounding, over-sensitivity to such grounding may also introduce bias and results in false positive.  

The model also exhibits bias for the group identifier term "Islamic" in Figure~\ref{fig:error-analysis}(b). Inspecting the text-visual alignments, we observed that the word "Islamic" does not assign much attention weights to any bounding regions.  We postulate that the bias for this group identifier term happens in the unimodal text space rather than the multimodal space.

\section{Conclusion}

We have presented an analysis of applying visual-linguistic models on the hateful meme classification task. Our analysis showed that the image modality contributes more to the hateful meme classification task, and the visual-linguistic models can perform visual-text slurs grounding. Nevertheless, the visual-linguistic models have also acquired biases, which resulted in false-positive predictions. For our future work, we will extend our analysis to multiple hateful meme datasets, and benchmark more models. We would also explore debiasing techniques to reduce the multimodal biases in the visual-linguistic models and improve their hateful meme classification performance.



\bibliographystyle{ACM-Reference-Format}
\balance
\bibliography{ref}

\end{document}